\documentclass[sigconf]{acmart}
\usepackage{multirow}
\usepackage{balance}

\usepackage{amssymb}
\usepackage{pifont}
\usepackage{caption}
\usepackage{subcaption}
\newcommand{\cmark}{\ding{51}}%
\newcommand{\xmark}{\ding{55}}%
\newcommand{\rulesep}{\unskip\ \vrule\ }
\begin{document}

\title{A Frustratingly Simple Approach for End-to-End\\Image Captioning}

\author{Ziyang Luo}
\affiliation{%
  \institution{Hong Kong Baptist University}
  \city{Hong Kong}
  \country{China}}
\email{cszyluo@comp.hkbu.edu.hk}

\author{Yadong Xi}
\affiliation{%
  \institution{Fuxi AI Lab, NetEase Inc.}
  \city{Hangzhou}
  \country{China}}
\email{xiyadong@corp.netease.com}

\author{Rongsheng Zhang}
\affiliation{%
  \institution{Fuxi AI Lab, NetEase Inc.}
  \city{Hangzhou}
  \country{China}}
\email{zhangrongsheng@corp.netease.com}

\author{Jing Ma}
\affiliation{%
  \institution{Hong Kong Baptist University}
  \city{Hong Kong}
  \country{China}}
\email{majing@hkbu.edu.hk}


\begin{abstract}
    Image Captioning is a fundamental task to join vision and language, concerning about cross-modal understanding and text generation.
    Recent years witness the emerging attention on image captioning.
    Most of existing works follow a traditional two-stage training paradigm. Before training the captioning models, an extra object detector is utilized to recognize the objects in the image at first. However, they require sizeable datasets with fine-grained object annotation for training the object detector, which is a daunting task.
    In addition, the errors of the object detectors are easy to propagate to the following captioning models, degenerating models' performance.
    To alleviate such defects, we propose a frustratingly simple but highly effective end-to-end image captioning framework, \textbf{Visual Conditioned GPT (VC-GPT)}, by connecting the pre-trained visual encoder (CLIP-ViT) and language decoder (GPT2). Different from the vanilla connection method that directly inserts the cross-attention modules into GPT2, we come up with a self-ensemble cross-modal fusion mechanism that comprehensively considers both the single- and cross-modal knowledge.
    As a result, we do not need extra object detectors for model training.
    Experimental results conducted on three popular image captioning benchmarks (MSCOCO, Flickr30k and NoCaps) demonstrate that our \textbf{VC-GPT} achieves either the best or the second-best performance across all evaluation metrics over extensive baseline systems.
\end{abstract}



\maketitle

\section{Introduction}

\begin{figure}
     \centering
    \begin{subfigure}[b]{0.5\textwidth}
       \includegraphics[height=3.2cm]{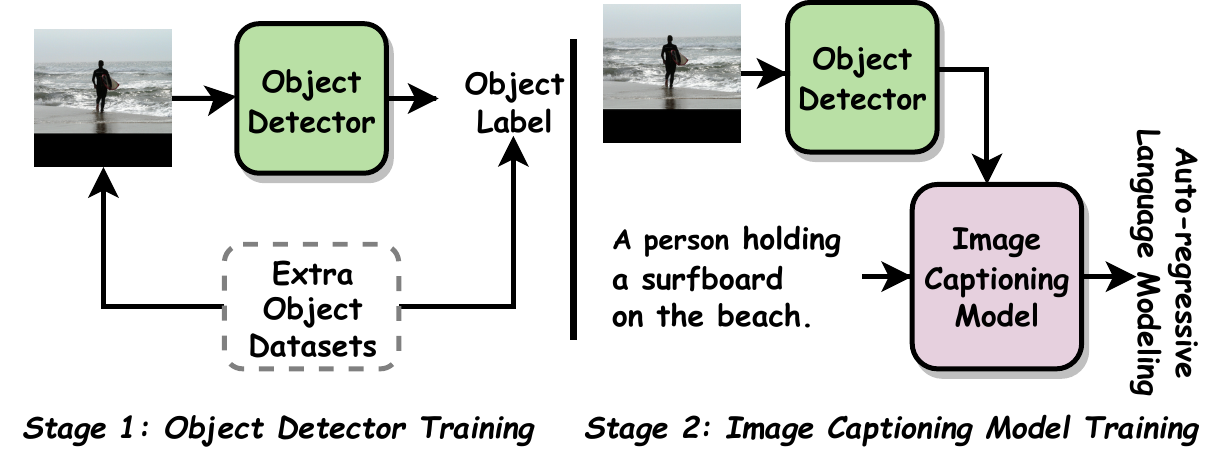}
       \caption{The two-stage training paradigm.}
       \label{fig:VCG2S} 
    \end{subfigure}
    
    \begin{subfigure}[b]{0.5\textwidth}
       \includegraphics[height=2.75cm]{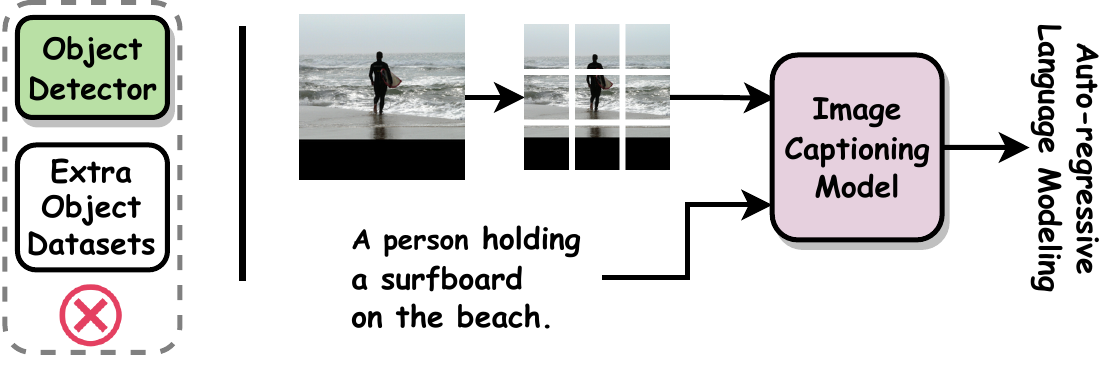}
       \caption{The end-to-end training paradigm.}
       \label{fig:VCGe2e}
    \end{subfigure}
    \caption{Comparing the end-to-end image captioning paradigm with the previous two-stage training paradigm.}
    \label{fig:VCG}
\end{figure}

Image Captioning is an important vision-and-language task to automatically generate description for an image with natural language~\cite{Stefanini2022FromST}, which involves visual recognition and language generation. Most existing image captioning approaches adopt a two-stage training paradigm~\cite{anderson2018bottomup,yao2018exploring,Yang2019AutoEncodingSG,huang2019attention,Cornia2019M2MT,Pan2020XLinearAN,li2020oscar,DBLP:conf/nlpcc/XiaHDZJSCBZ21} as illustrated in Figure~\ref{fig:VCG2S}. 
More specifically, an object detector (OD) is trained with extra fine-grained object recognition datasets to detect the visual objects in an image~\cite{NIPS2015_14bfa6bb} at first. Then image captioning models are trained on top of the detected object with feature learning algorithms. However, the drawback of such paradigm is that training the ODs needs large-scale and high-quality object recognition datasets which are time-consuming to annotate~\cite{SALARI2022} and only available for popular datasets, like MSCOCO~\cite{lin2015microsoft}. In addition, the errors of ODs are easy to propagate to the following captioning models.

\begin{figure*}[h!]
     \centering
     \begin{subfigure}[b]{0.25\textwidth}
         \centering
         \includegraphics[width=4cm]{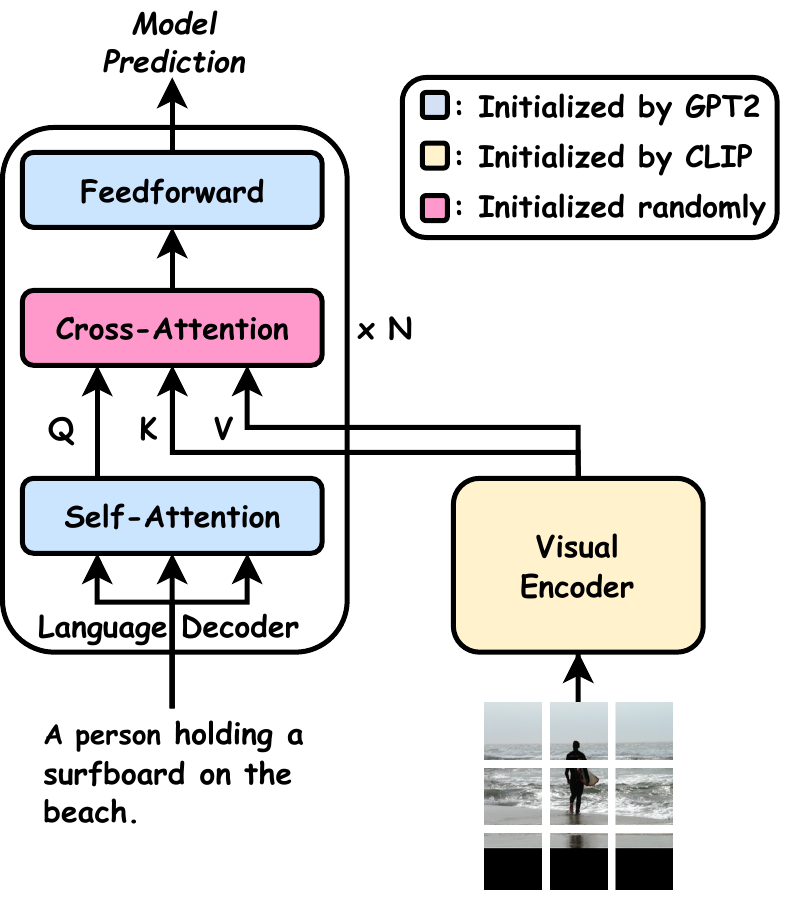}
         \caption{The Vanilla Framework.}
         \label{fig:vcg_features_vanilla}
     \end{subfigure}
     \rulesep
     \begin{subfigure}[b]{0.7\textwidth}
         \centering
         \includegraphics[width=12cm]{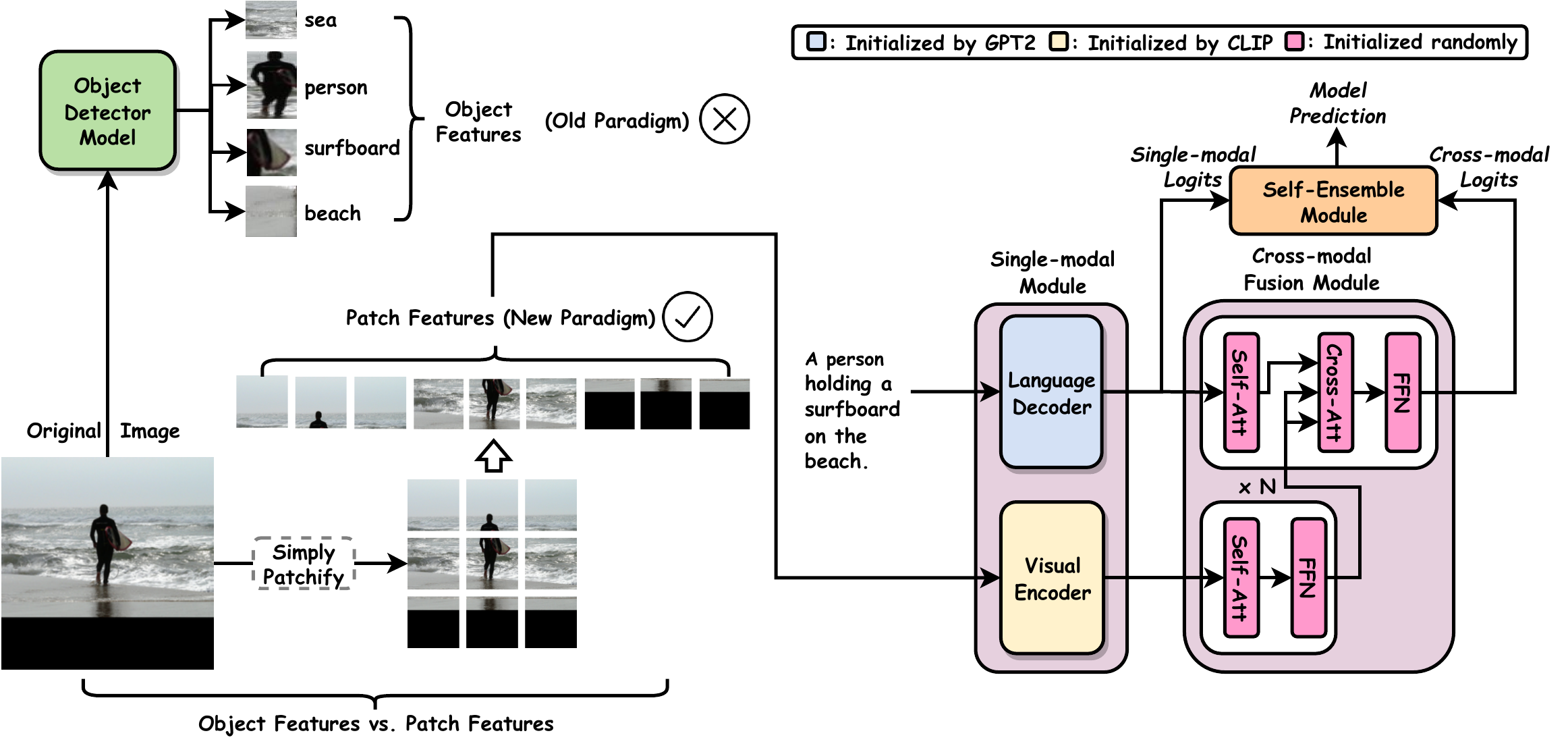}
         \caption{Our End-to-End VC-GPT Framework.}
         \label{fig:vcg_features_our}
     \end{subfigure}
        \caption{Comparing the differences between the Vanilla Framework and our \textbf{VC-GPT} Framework.}
        \label{fig:vcg_features}
\end{figure*}
To remove the object detectors from the captioning models training pipeline, the natural method is to leverage the pre-trained vision transformer~\cite{bao2021beit,he2021masked,zhou2021ibot,dosovitskiy2021image} and directly separate an image into several patches which are then fed into the model. Since these models are pre-trained with extremely large-scale datasets, they can generalize to a wide range of natural images. Thus, there is no need for additional object annotations and the captioning models can be trained in an end-to-end manner. Figure~\ref{fig:VCGe2e} illustrates such paradigm. In our work, we follow this paradigm and adopt the state-of-the-art pre-trained vision transformer, CLIP-ViT~\cite{radford2021learning} as the visual encoder and language transformer, GPT2~\cite{radford2019language} as the language decoder. Due to the superior language generation power of GPT2 and the visual recognition ability of CLIP-ViT, this method can reduce the burden of learning these two abilities and focus on aligning the cross-modal information during training.

Unfortunately, GPT2 lacks an essential cross-attention component to process the information from the encoder. To fill up this gap, the simplest method is to directly insert the randomly initialized cross-attention modules into GPT2, evolving to a standard seq2seq encoder-decoder Transformer~\cite{NIPS2017_3f5ee243}. Figure~\ref{fig:vcg_features_vanilla} introduces such vanilla framework. Since there exists a gap between language and vision modals, directly injecting the vision information into the internal layer of GPT2 is highly possible to harm the pre-trained language knowledge of GPT2. Therefore, it is necessary for us to come up with a more suitable method to connect CLIP-ViT and GPT2 for end-to-end image captioning.

Inspired by the encoder-decoder fusion methods in neural machine translation~\cite{rothe-etal-2020-leveraging,ma2020xlmt,guo2020incorporating,sun-etal-2021-multilingual-translation}, we introduce a frustratingly simple but highly effective self-ensemble cross-modal fusion mechanism that disentangles the single- and cross-modal knowledge. Figure~\ref{fig:vcg_features_our} depicts our method. The language decoder and the fusion module serve as the single- and cross-modal language generation experts, respectively. The self-ensemble module combines their predictions, comprehensively considering both the single- and cross-modal information and leading to better performance. Based on such design, we propose our \textbf{Visual Conditioned GPT (VC-GPT)} framework for end-to-end image captioning. Evaluating on three popular image captioning benchmarks (MSCOCO~\cite{lin2015microsoft}, Flickr30k~\cite{plummer2016flickr30k} and NoCaps~\cite{Agrawal2019nocapsNO}), our \textbf{VC-GPT} achieves either the best or the second-best performance across all evaluation metrics over extensive baseline systems. Beyond this, we also conduct extra empirical exploration on how different training settings affect our model performance.

Our contributions can be summarized as follows:
\begin{itemize}
    \item We propose an end-to-end image captioning framework, \textbf{Visual Conditioned GPT (VC-GPT)}, removing the need of extra object detectors.
    \item We introduce a frustratingly simple but highly effective \textbf{self-ensemble mechanism} that comprehensively considers the single- and cross-modal knowledge for natural language generation, leading to much better performance. (section~\ref{sec:cross} and \ref{sec:vanilla})
    \item Comparing with the end-to-end image captioning baseline systems, our \textbf{VC-GPT} outperforms all of them over all evaluation metrics. (section~\ref{sec:e2e})
    \item Comparing with the traditional two-stage training baseline systems, our \textbf{VC-GPT} achieves the best or the second-best performance across all evaluation metrics. (section~\ref{sec:2stage})
\end{itemize}

\section{Related Work}

\noindent\textbf{CLIP-ViT and GPT2.}\indent Recently, \citet{radford2021learning} propose a novel model, CLIP, which contains two encoders, one for images and one for texts. Such models are pre-trained with contrastive loss~\cite{He2020MomentumCF} to jointly represent the image and its paired caption in the same semantic latent space. This model has already achieved great success in many computer vision or cross-modal tasks, like image classification~\cite{Zhou2021LearningTP} and text-to-image generating/editing~\cite{Bau2021PaintBW,Gal2021StyleGANNADACD,Zhou2021LAFITETL}. In this work, we adapt the powerful CLIP model to the image-to-text generation scenario. Since there is no input text, we only employ the image encoder (CLIP-ViT) in CLIP to represent an image. CLIP-ViT follows the same settings as the previous vision transformer (ViT~\cite{dosovitskiy2021image}) to directly divide an image into patch features, so that we can get rid of the object detectors.

In addition, GPT2~\cite{radford2019language} is a pre-trained transformer decoder-only language model, which is widely used in the natural language processing area, including text summarization~\cite{Khandelwal2019SampleET,Yang2021ResearchOA} and neural machine translation~\cite{rothe-etal-2020-leveraging,ma2020xlmt,guo2020incorporating,sun-etal-2021-multilingual-translation}. In our work, we propose a simple, yet effective \textbf{VC-GPT} framework to leverage the powerful visiual recognition and language generation abilities of CLIP-ViT and GPT2 for caption generation.\\

\noindent\textbf{Image Captioning.}\indent Generating the language descriptions from images is an important task to examine the vision-and-language representation ability of a cross-modal model~\cite{10.1145/3295748}. The pioneer works for image captioning are majorly template-based~\cite{5540112,5487377}, which generate slotted templates and use the detected objects to fill in the slots. In recent years, the neural-based methods become popular, utilizing the encoder-decoder framework~\cite{Sutskever2014SequenceTS} to generate syntactically and semantically correct sentences given images. Early works are based on the features extracted by a pre-trained image classification model~\cite{Chen2017SCACNNSA,Chen2014LearningAR}, while the later works~\cite{rennie2017selfcritical,lu2017knowing,anderson2018bottomup,lu2018neural,yao2018exploring,zhou2019grounded,huang2019attention,cornia2020meshedmemory,pan2020xlinear} choose to exploit the object detectors to extract the visual features. However, the pre-trained object detectors are only available for the popular benchmarks, like MSCOCO~\cite{lin2015microsoft} and Flickr30k~\cite{plummer2016flickr30k}. For the new datasets, most previous methods require additional object detection annotations. To remedy such defects, our work chooses to adopt the CLIP-ViT to extract the visual features. Since it is trained over an extremely large-scale images dataset, it can be applied to a wide range of natural images without additional annotations.\\

\noindent\textbf{Vision-and-Language Pre-training.}\indent Since 2019, a large amount of works~\cite{li2019visualbert,tan2019lxmert,li2019unicodervl,chen2020uniter,Su2020VL-BERT} propose vision-and-language pre-trained models to process the vision and text information within a single model. For example, VL-T5~\cite{pmlr-v139-cho21a} and XGPT~\cite{DBLP:conf/nlpcc/XiaHDZJSCBZ21} use visual tokens extracted by the object detector as a prefix to text tokens. Then the entire models are pre-trained in an auto-regressive manner. OSCAR~\cite{li2020oscar} utilizes the same structure as BERT and requires the additional supervision of object tags. Hence, the object detector is an indispensable component of these works. To mitigate the need for object detectors, METER~\cite{dou2021empirical} directly connects RoBERTa~\cite{Liu2019RoBERTaAR} and CLIP-ViT in a single model. Different from ours, METER majorly focuses on discriminative cross-modal tasks, like Visual Question Answering~\cite{agrawal2016vqa} and Visual Language Inference~\cite{xie2019visual}. Our work follows the similar idea of METER to connect GPT2 and CLIP-ViT with our self-ensemble fusion mechanism for the generative cross-modal task, image captioning.

\section{Methodology}

\noindent\textbf{Overview.}\indent To remove the need of object detectors and train the image captioning system in an end-to-end manner, we leverage the pre-trained visual transformer (CLIP-ViT) as encoder and language transformer (GPT2) as decoder, making up the basic single-modal module. Since they only can process the vision- and language-only information, a cross-modal fusion module is an inevitable component to connect them. In section~\ref{sec:cross}, we present our self-ensemble cross-modal fusion mechanism to fill up such gap.

\subsection{Basic Single-Modal Module}\label{sec:basic}

\noindent\textbf{Visual Encoder.}\indent We follow the same setting as the visual transformer (ViT) which recently attracts much attention and shows remarkable performance in the computer vision area~\cite{dosovitskiy2021image,radford2021learning,bao2021beit,zhou2021ibot,he2021masked,zhou2022image}. ViT does not need the help of the extra object detectors, but directly takes a sequence of image patches as input and visual representation hidden states for each patch as output. In our framework, the visual encoder is initialized by the state-of-the-art pre-trained ViT, \textbf{CLIP-ViT}, which achieves superior performance across various visual benchmarks, like ImageNet~\cite{imagenet_cvpr09}. Our framework can benefit from the excellent visual representation ability of such model.\\

\noindent\textbf{Language Decoder.}\indent The single-modal language decoder is initialized by the state-of-the-art language generation pre-trained model, \textbf{GPT2}. Most of the previous works neglect the importance of single-modal generation ability, but train their language decoder from scratch~\cite{DBLP:conf/nlpcc/XiaHDZJSCBZ21,pmlr-v139-cho21a,fang2021injecting}. As a result, their models need to spend extra effort on learning how to model language and generate smooth sentences, which increases the burden on their models. Since GPT2 has remarkable language generation ability, this design can alleviate the workload of the model during cross-modal training. As a result, the model can focus on aligning the cross-modal information.

\subsection{Cross-Modal Fusion Module}\label{sec:cross}

In this part, we first discuss the vanilla framework and its drawbacks in connecting CLIP-ViT and GPT2. Then we present our \textbf{VC-GPT} framework to avoid its problems and propose our self-ensemble mechanism to further improve the performance.\\

\noindent\textbf{The Vanilla Framework.}\indent Figure~\ref{fig:vcg_features_vanilla} shows the structure of the vanilla framework to connect CLIP-ViT and GPT2 for image captioning by directly inserting the randomly initialized cross-attention modules into GPT2. With such design, GPT2 evolves from a single-modal to a cross-modal language generation model. One can consider this as a special case of Continual Lifelong Learning~\cite{Parisi2018ContinualLL} that GPT2 continually learns the cross-modal knowledge. However, there exists a modal gap between language and vision. Directly injecting the visual information into GPT2 is highly possible to cause the catastrophic forgetting problem~\cite{Kaushik2021UnderstandingCF}, harming the original language generation knowledge and degenerating captioning performance.\\

\noindent\textbf{Our VC-GPT Framework.}\indent Different from the vanilla framework, we do not insert any randomly initialized parameters into our single-modal module, but the visual encoder is followed by several extra encoder layers and the language decoder is followed by several extra decoder layers with cross-attention. As a result, we can maintain the integrality of GPT2 and minimize the negative influence on the single-modal module.

\begin{figure}
    \centering
    \includegraphics[height=6cm]{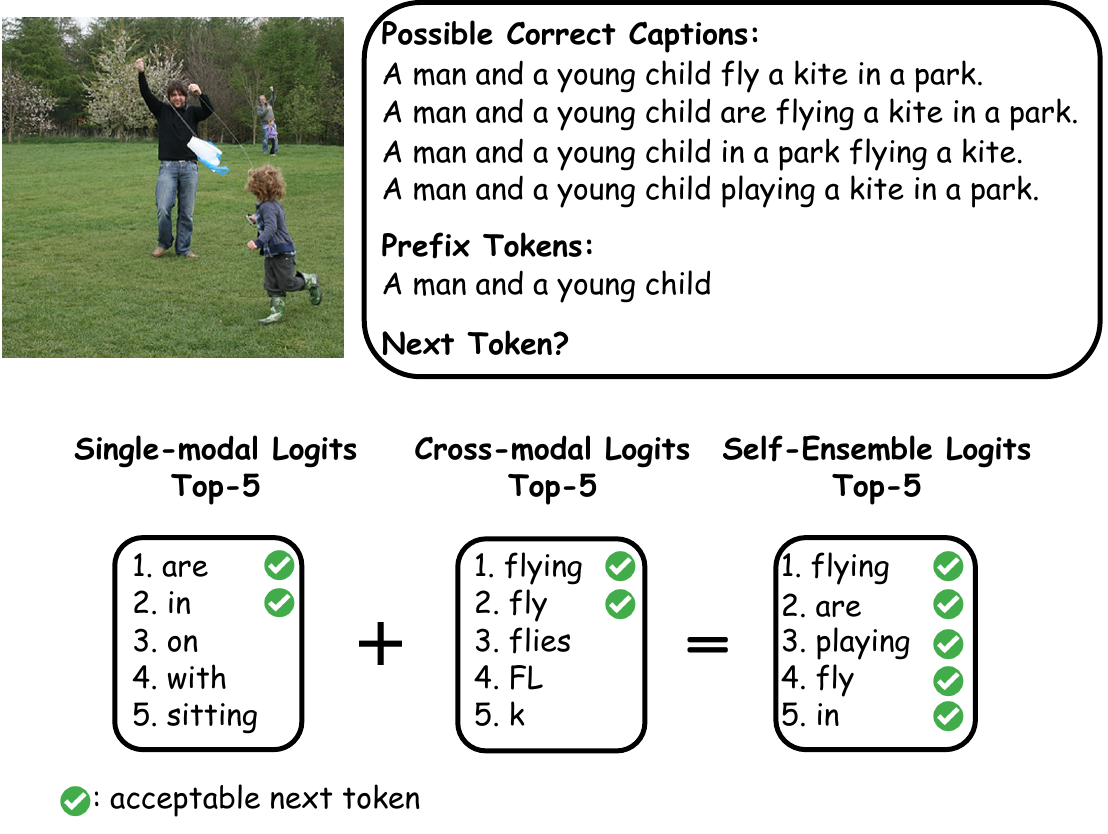}
    \caption{Self-Ensemble mechanism can comprehensively consider both single- and cross-modal knowledge, leading to better performance. All logits are generated by our VC-GPT.}
    \label{fig:selfensemble}
\end{figure}

After connecting the vision and language modals, one would directly adopt the cross-modal representation for caption generation. However, such method neglects that the GPT2-initialized language decoder also has superior language generation ability. The single-modal predictions can provide a language-only generation perspective. By combing them with the cross-modal predictions, we can generate more smooth sentences. To achieve our goal, we introduce a frustratingly simple, but effective \textbf{self-ensemble} module to combine the output logits of the cross-modal fusion module and the single-modal language decoder with a residual connection~\cite{he2015deep}:
\begin{equation}\label{eq:residual}
    logits=W^{G}h^{G} + W^{Fuse}h^{Fuse},
\end{equation}
where $W^{G}$ is a linear projection of language decoder, initialized by the word embeddings matrix of GPT2. $W^{Fuse}$ is a linear projection of the fusion module and initialized randomly. Figure~\ref{fig:selfensemble} shows an example which is generated by our \textbf{VC-GPT}. One can find that only the top-2 tokens are acceptable for the single- and cross-modal logits. After combing them with self-ensemble, all the top-5 tokens are acceptable. Therefore, the self-ensemble mechanism can comprehensively consider both single-modal and cross-modal knowledge, leading to better performance.

\subsection{Training Objective}\label{sec:objective}

During training, the parallel image-caption data are needed for our \textbf{VC-GPT} to learn how to predict the next token with the past tokens and the visual information. The pre-training and fine-tuning objective is the auto-regressive language modeling conditioned on the visual information:
\begin{equation}
    \mathcal{L}=-\Sigma_{t=1}^T\log P(x_t|x_{<t}, V_{info}),
\end{equation}
where $V_{info}$ represents the visual information of the image encoded by the visual encoder, T denotes the length of a sequence and $x_{<t} = (x_0,...,x_{t-1})$. The probability of the current token in the $t^{th}$ position is determined by all the past tokens and the visual information, $V_{info}$. We can also denote this objective as cross-entropy optimization.

Apart from this basic training strategy, most previous works~\cite{yao2018exploring,anderson2018bottomup,Cornia2019M2MT,pan2020xlinear,li2020oscar} also indicate that after optimizing with cross-entropy loss, continually training image captioning systems with REINFORCE algorithms~\cite{NIPS1999_464d828b} can boost the performance by a large margin. We follow this paradigm to directly optimize the image captioning evaluation metric (CIDEr~\cite{vedantam2015cider}) with the widely used self-critical sequence training (SCST) REINFORCE algorithm~\cite{Rennie2017SelfCriticalST}. We denote this training strategy as CIDEr optimization.

\begin{table}
    \small
    \centering
    \caption{Comparing with the end-to-end image captioning baseline models on MSCOCO test set. Models are based on cross-entropy loss. ``-'' represents that the original paper does not report this result. ``\#Images'' represents the number of distinct images during captioning model pre-training.}
    \begin{tabular}{l|c|c|cccc}
        \hline
        \multirow{2}{*}{\textbf{Model}} & \multirow{2}{*}{\textbf{\#Images}} & \multirow{2}{*}{\textbf{OD?}} & \multicolumn{4}{c}{\textbf{MSCOCO (test)}}\\
        & & & \textbf{C} & \textbf{B@4} & \textbf{M} & \textbf{S}\\
        \hline
        \hline
        \multicolumn{6}{l}{\textit{End-to-End Image Captioning Baseline Systems}}\\
        \hline
        \textbf{ViLT-CAP}~\cite{fang2021injecting} & 4M & \xmark & 113.5 & 33.7 & 27.7 & 20.9\\
        \textbf{ViTCAP}~\cite{fang2021injecting} & 4M & \xmark & 125.2 & 36.3 & 29.3 & 22.6\\
        \textbf{E2E-VLP}~\cite{xu-etal-2021-e2e} & 180k & \xmark & 117.3 & 36.2 & - & -\\
        \textbf{I-Tuning}~\cite{luo2022ituning} & 110k & \xmark & 122.1 & 36.1 & 29.4 & 22.6\\
        \hline
        \hline
         \multicolumn{6}{l}{\textit{Our Models}}\\
        \hline
        \textbf{Vanilla} & 110k & \xmark & 120.6 & 35.5 & 29.4 & 22.7\\
        \textbf{VC-GPT w/o SE} & 110k & \xmark & 120.0 & 34.3 & 30.0 & 23.4\\
        \textbf{VC-GPT} & 110k & \xmark & \textbf{127.7} & \textbf{37.3} & \textbf{30.1} & \textbf{23.8}\\
        \hline
        \multicolumn{6}{l}{*(\textit{SE = self-ensemble, OD = object detector.})}\\
    \end{tabular}
    \label{tab:E2E}
\end{table}

\begin{table*}[h!]
    \centering
    \caption{Comparing with the two-stage training baseline models on MSCOCO. \textbf{Bold} indicates the best score. \underline{Underline} indicates the second best score. ``\#Images'' represents the number of distinct images during captioning model pre-training. OD corresponds to object detector. ``-'' represents that the original paper does not report this result.}
    \begin{tabular}{l|c|c|cccc|cccc}
        \hline
        \multicolumn{10}{c}{\textbf{MSCOCO Image Captioning (Karpathy's test split)}}\\
        \hline
         \multirow{2}{*}{\textbf{Model}} & \multirow{2}{*}{\textbf{OD?}} & \multirow{2}{*}{\textbf{\#Images}} & \multicolumn{4}{c|}{\textbf{Cross Entropy Optimization}} & \multicolumn{4}{c}{\textbf{CIDEr Optimization}}\\
        & & & \textbf{CIDEr} & \textbf{BLEU@4} & \textbf{METEOR} & \textbf{SPICE} & \textbf{CIDEr} & \textbf{BLEU@4} & \textbf{METEOR} & \textbf{SPICE}\\
        \hline
        \hline
        \multicolumn{9}{l}{\textit{Do not use any vision-and-language pre-trained models}}\\
        \hline
        \textbf{BUTD}~\cite{anderson2018bottomup} & \cmark & 0 & 113.5 & 36.2 & 27.0 & 20.3 & 120.1 & 36.3 & 27.7 & 21.4\\
        \textbf{LBPF}~\cite{yao2018exploring} & \cmark & 0 & 116.4 & 37.4 & 28.1 & 21.2 & 127.6 & 38.3 & 28.5 & 22.0\\
        \textbf{SGAE}~\cite{Yang2019AutoEncodingSG} & \cmark & 0 & 116.7 & 36.9 & 27.7 & 20.9 & 127.8 & 38.4 & 28.4 & 22.1\\
        \textbf{AoANet}~\cite{huang2019attention} & \cmark & 0 & 119.8 & 37.2 & 28.4 & 21.3 & 129.8 & 38.9 & 29.3 & 22.4\\
        \textbf{M$^2$ Transformer}~\cite{Cornia2019M2MT} & \cmark & 0 & - & - & - & - & 131.2 & 39.1 & 29.2 & 22.6\\
        \textbf{X-LAN}~\cite{Pan2020XLinearAN} & \cmark & 0 & 122.0 & \textbf{38.2} & 28.8 & 21.9 & 132.0 & 39.5 & 29.5 & 23.4\\
        \hline
        \hline
        \multicolumn{9}{l}{\textit{Large-scale cross-modal pre-trained models}}\\
        \hline
        \textbf{MiniVLM}~\cite{Wang2020MiniVLMAS} & \cmark & 14M & 119.8 & 35.6 & 28.6 & 21.6 & 131.7 & 39.2 & \underline{29.7} & \underline{23.5}\\
        \textbf{OSCAR$_{b}$}~\cite{li2020oscar} & \cmark & 4M & \underline{123.7} & 36.5 & \textbf{30.3} & \underline{23.1} & \textbf{137.6} & \underline{40.5} & \underline{29.7} & 22.8\\
        \textbf{UniVL}~\cite{liu2021unified} & \cmark & 3M & 116.8 & 35.6 & 28.6 & 21.4 & - & - & - & -\\
        \textbf{XGPT}~\cite{DBLP:conf/nlpcc/XiaHDZJSCBZ21} & \cmark & 3M & 120.1 & 37.2 & 28.6 & 21.8 & - & - & - & -\\
        \textbf{UVLP}~\cite{zhou2019unified} & \cmark & 3M & 116.9 & 36.5 & 28.4 & 21.2 & 129.3 & 39.5 & 29.3 & 23.2\\
        \hline
        \multicolumn{9}{l}{\textit{Small-scale cross-modal pre-trained models}}\\
        \hline
        \textbf{UNICORN}~\cite{yang2021crossing} & \cmark & 200k & 119.1 & 35.8 & 28.4 & 21.5 & - & - & - & -\\
        \textbf{VL-T5}~\cite{pmlr-v139-cho21a} & \cmark & 180k & 116.5 & 34.5 & 28.7 & 21.9 & - & - & - & -\\
        \hline
        \hline
        \multicolumn{9}{l}{\textit{Our Models}}\\
        \hline
        \textbf{VC-GPT} & \xmark & \textbf{110k} & \textbf{127.7} & \underline{37.3} & \underline{30.1} & \textbf{23.8} & \underline{137.5} & \textbf{41.4} & \textbf{30.0} & \textbf{24.0}\\
        \hline
    \end{tabular}
    \label{tab:previous_work_comparsion}
\end{table*}

\begin{table}
    \centering
    \caption{Results on Flickr30k Image Captioning. \textbf{Bold} indicates the best score. \underline{Underline} indicates the second best score. For a cross-modal pre-trained model, the number in the column of ``\#Images'' represents the number of distinct images during captioning model pre-training.}
    \begin{tabular}{l|c|cccc}
        \hline
        \multicolumn{6}{c}{\textbf{Flickr30k Image Captioning (Karpathy's test split)}}\\
        \hline
        \multirow{2}{*}{\textbf{Model}} & \multirow{2}{*}{\textbf{\#Images}} & \multicolumn{4}{c}{\textbf{Cross Entropy Optimization}}\\
        & & \textbf{C} & \textbf{B@4} & \textbf{M} & \textbf{S}\\
        \hline
        \hline
        \multicolumn{6}{l}{\textit{Baseline Systems}}\\
        \hline
        \textbf{BUTD}~\cite{anderson2018bottomup} & 0 & 56.6 & 27.3 & 21.7 & 16.0\\
        \textbf{NBT}~\cite{lu2018neural} & 0 & 57.5 & 27.1 & 21.7 & 15.6\\
        \textbf{GVD}~\cite{zhou2019grounded} & 0 & 62.3 & 27.3 & 22.5 & 16.5\\
        \textbf{UVLP}~\cite{zhou2019unified} & 3M & 67.4 & 30.1 & 23.0 & 17.0\\
        \textbf{XGPT}~\cite{DBLP:conf/nlpcc/XiaHDZJSCBZ21} & 3M & \underline{70.9} & \textbf{31.8} & 23.6 & \underline{17.6}\\
        \textbf{UNICORN}~\cite{yang2021crossing} & 200k & 70.1 & 30.7 & \underline{23.7} & 17.4\\
        \hline
        \hline
         \multicolumn{6}{l}{\textit{Our Models}}\\
        \hline
        \textbf{VC-GPT} & 110k & \textbf{79.6} & \underline{31.2} & \textbf{25.8} & \textbf{20.1}\\
        \hline
    \end{tabular}
    \label{tab:flickr}
\end{table}

\section{Experiment Setup}\label{sec:setup}

\noindent\textbf{Default Model Settings.}\indent Our framework adopts the CLIP-ViT B/16\footnote{\url{https://huggingface.co/openai/clip-vit-base-patch16}} pre-trained model as our visual encoder and GPT2-base\footnote{\url{https://huggingface.co/gpt2}} pre-trained model as our language decoder. The visual encoder and language decoder are followed by 6 extra fusion layers. The original parameters of the visual encoder (CLIP) are frozen during pre-training and updated during fine-tuning. All parameters in the extra cross-modal fusion module are initialized randomly.\\

\noindent\textbf{Datasets.}\indent For pre-training, we leverage a popular dataset, Visual Genome\footnote{\url{http://visualgenome.org/}} which is commonly used in the previous works. Such dataset contains 110k distinct images and 5.4M short region descriptions. For image captioning task, we use three different datasets, including \textbf{MSCOCO Captions}~\citep{lin2015microsoft}\footnote{\url{https://cocodataset.org/\#home}}, \textbf{Flickr30k}~\citep{plummer2016flickr30k}\footnote{\url{http://hockenmaier.cs.illinois.edu/DenotationGraph/}} and \textbf{NoCaps}~\citep{Agrawal2019nocapsNO}.\footnote{\url{https://nocaps.org/}} For \textbf{MSCOCO} and \textbf{Flickr30k}, each image corresponds to around 5 captions. We follow the standard Karpathy’s split~\cite{karpathy2015deep} to split 113.2k/5k/5k and 29.8k/1k/1k images for train/val/test, respectively. For \textbf{NoCaps}, it measures generalization to unseen object classes, containing only validation and test sets. Models are only trained with the same dataset as MSCOCO. We follow the suggestion of \citet{li2020oscar} to evaluate our models with the validation set, containing 4.5k images and 10 captions per image.\\

\noindent\textbf{Evaluation Metrics.}\indent We use 4 standard automatic evaluation metrics for image captioning, including CIDEr~\cite{vedantam2015cider}, BLEU@4~\cite{papineni-etal-2002-bleu}, METEOR~\cite{banerjee-lavie-2005-meteor} and SPICE~\cite{anderson2016spice}. For simplicity, we also use C to denote CIDEr, B@4 to denote BLEU@4, M to denote METEOR and S to denote SPICE. For MSCOCO and Flickr30k, the scores are calculated by the official code.\footnote{\url{https://github.com/tylin/coco-caption}} For NoCaps, we evaluate our models with the official online server.\footnote{\url{https://eval.ai/web/challenges/challenge-page/355/overview}}\\

\noindent\textbf{Baseline Systems.}\indent To obtain a comprehensive comparison with the previous works, we include three categories of baseline systems:

(1) \textbf{Models w/o Cross-modal Pre-training}: we include 9 works for this type, including BUTD~\cite{anderson2018bottomup}, NBT~\cite{lu2018neural}, LBPF~\cite{yao2018exploring}, GVD~\cite{zhou2019grounded}, SGAE~\cite{Yang2019AutoEncodingSG}, AoANet~\cite{huang2019attention}, M$^2$ Transformer~\cite{Cornia2019M2MT}, X-LAN~\cite{pan2020xlinear} and ClipCap~\cite{mokady2021clipcap}. All of them achieve superior results without any cross-modal pre-training;

(2) \textbf{Large-scale cross-modal pre-trained models}: we include the results of 7 cross-modal pre-trained models, including UVLP~\cite{zhou2019unified}, OSCAR~\cite{li2020oscar}, XGPT~\cite{DBLP:conf/nlpcc/XiaHDZJSCBZ21}, MiniVLM~\cite{Wang2020MiniVLMAS}, ViLT-CAP~\cite{fang2021injecting}, ViTCAP~\cite{fang2021injecting} and UniVL~\cite{liu2021unified}. All of them consume large-scale distinct images during cross-modal pre-training;

(3) \textbf{Small-scale cross-modal pre-trained models}: we include the results of 4 cross-modal pre-trained models, E2E-VLP~\cite{xu-etal-2021-e2e}, UNICORN~\cite{yang2021crossing}, VL-T5~\cite{pmlr-v139-cho21a} and I-Tuning~\cite{luo2022ituning}. These models consume the same-level parallel data as ours.\\

\noindent\textbf{Implementation Details.}\indent For all cross-entropy-based experiments, we train our models with the AdamW optimization algorithm~\cite{loshchilov2019decoupled}, 4k batch size, mixed-precision training and FP16. We pre-train our models for 10 epochs and fine-tune our models for 30 epochs. The warm-up step is set to 10\% of the total training steps and the learning rate decay strategy is linear. We use center-crop to resize each image into the size of 288x288 for pre-training and 480x480 for fine-tuning. Beyond this, we use the beam search (beam size = 5) to generate caption during evaluation. The detailed hyper-parameters can be found in the Supplemental Material. All experiments are conducted on 8 NVIDIA A100 GPUs.

\begin{table*}
    \centering
     \caption{Comparisons with previous models on NoCaps image captioning. \textbf{Bold} indicates the best scores.}
    \begin{tabular}{l|cc|cc|cc|cc}
        \hline
        \multicolumn{9}{c}{\textbf{NoCaps Image Captioning}}\\
        \hline
        \multicolumn{1}{l|}{\multirow{2}{*}{\textbf{Model}}} & \multicolumn{2}{c|}{\textbf{in-domain}} & \multicolumn{2}{c|}{\textbf{near-domain}} & \multicolumn{2}{c|}{\textbf{out-of-domain}} & \multicolumn{2}{c}{\textbf{Overall}}     \\
        \multicolumn{1}{l|}{} & \textbf{CIDEr} & \multicolumn{1}{c|}{\textbf{SPICE}} & \textbf{CIDEr} & \multicolumn{1}{c|}{\textbf{SPICE}} & \textbf{CIDEr} & \multicolumn{1}{c|}{\textbf{SPICE}} & \textbf{CIDEr} & \multicolumn{1}{c}{\textbf{SPICE}}\\
        \hline
        \hline
        \multicolumn{9}{l}{\textit{Baseline Systems}}\\
        \hline
        \textbf{Updown}~\cite{Agrawal2019nocapsNO} & 78.1 & 11.6 & 57.7 & 10.3 & 31.3 & 8.3 & 55.3 & 10.1\\
        \textbf{BUTD}~\cite{anderson2018bottomup} & 74.3 & 11.5 & 56.9 & 10.3 & 30.1 & 8.1 & 54.3 & 10.1\\
        \textbf{ClipCap}~\cite{mokady2021clipcap} & 84.9 & 12.1 & 66.8 & 10.9 & \textbf{49.1} & 9.6 & 65.8 & 10.9\\
        \textbf{OSCAR}$_{b}$~\cite{li2020oscar} & 79.6 & 12.3 & 66.1 & 11.5 & 45.3 & \textbf{9.7} & 63.8 & 11.2\\
        \hline
        \hline
        \multicolumn{9}{l}{\textit{Our Models}}\\
        \hline
        \textbf{VC-GPT} & \textbf{85.1} & \textbf{13.2} & \textbf{71.9} & \textbf{12.3} & 47.4 & \textbf{9.7} & \textbf{68.8} & \textbf{12.0}\\
        \hline
    \end{tabular}
    \label{tab:nocaps}
\end{table*}

\begin{figure*}
    \centering
    \includegraphics[height=8cm]{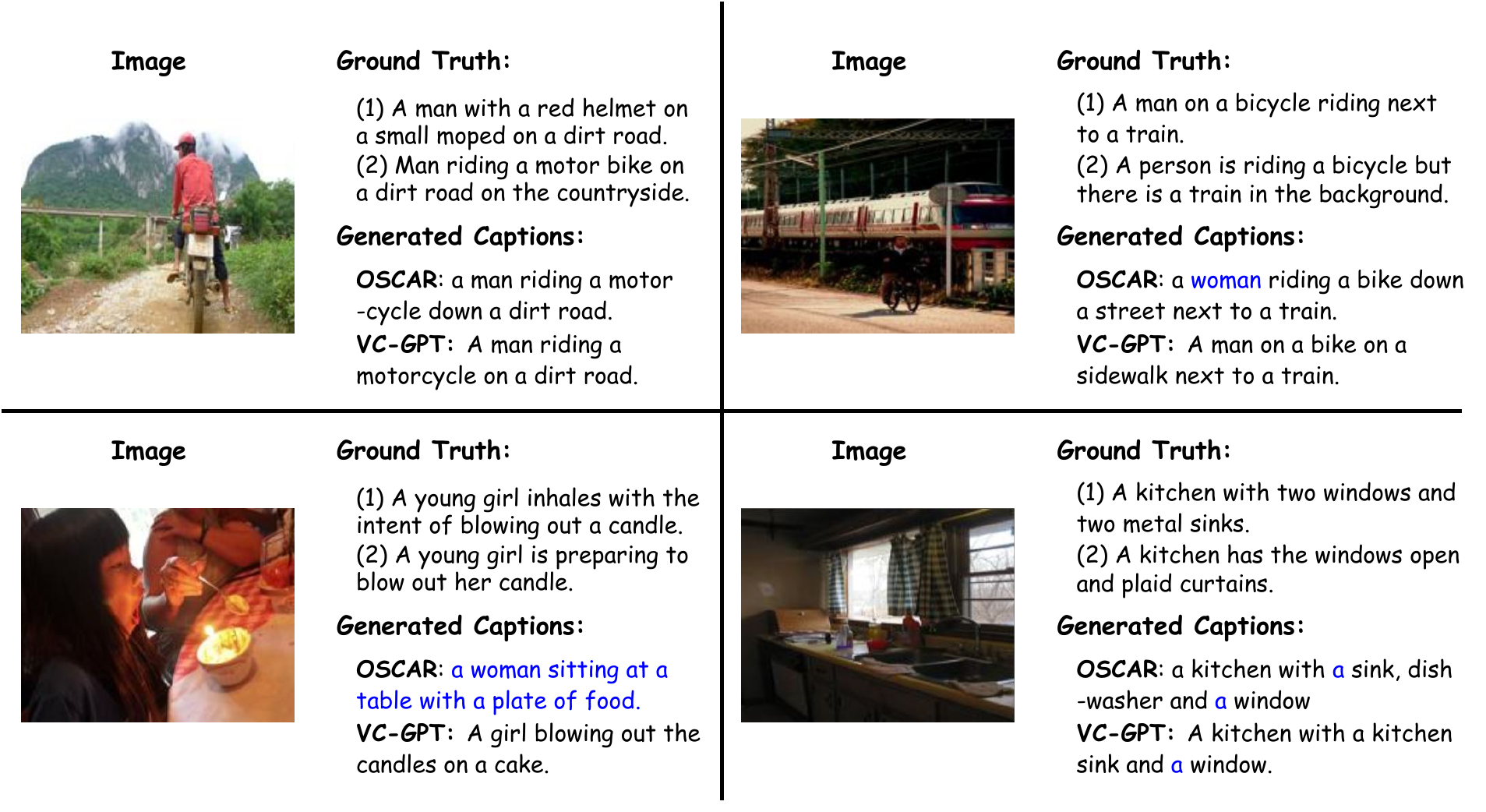}
    \caption{Image captioning examples of our VC-GPT and OSCAR for the first 4 images in the MSCOCO test set. {\color{blue} Blue} indicates that such caption or word is inaccurate, after comparing with the ground truth. For simplicity, we only show the first two golden captions per image.}
    \label{fig:generated_example}
\end{figure*}

\section{Experiment Results}\label{sec:exploration}

In this section, we present the image captioning experimental results on three popular benchmarks to examine the effectiveness of our \textbf{VC-GPT} framework. Especially, (1) we compare with the vanilla framework and verify the importance of our self-ensemble module in section~\ref{sec:vanilla}; (2) we compare with the end-to-end baseline systems in section~\ref{sec:e2e}; (3) we compare with the two-stage training baseline systems in section~\ref{sec:2stage}. (4) we evaluate our \textbf{VC-GPT} qualitatively in section~\ref{sec:qualitative}; (5) we present the cross-attention map example in section~\ref{sec:attention}.

\subsection{Comparing with Vanilla Framework}\label{sec:vanilla}

In section~\ref{sec:cross} and Figure~\ref{fig:vcg_features}, we have compared our fusion method with the vanilla framework which directly inserts the randomly initialized cross-attention modules into GPT2. Table~\ref{tab:E2E} shows that our method outperforms the vanilla method by a large margin (+7.1 CIDEr score), corroborating our claim that our design can generate more smooth sentences. 

\underline{\textbf{Is the Self-Ensemble helpful?}} We introduce a self-ensemble module in section~\ref{sec:cross} to combine the single- and cross-modal logits (see equation~\ref{eq:residual}). Table~\ref{tab:E2E} examines whether such module is effective. We can find that dropping such module degenerates our \textbf{VC-GPT} performance on all evaluation metrics. Especially, the CIDEr score decreases by more than 7 points. This result corroborates our claim in section~\ref{sec:cross} that the self-ensemble mechanism can lead to much better performance.

\begin{figure*}[h!]
    \centering
    \includegraphics[height=7.8cm]{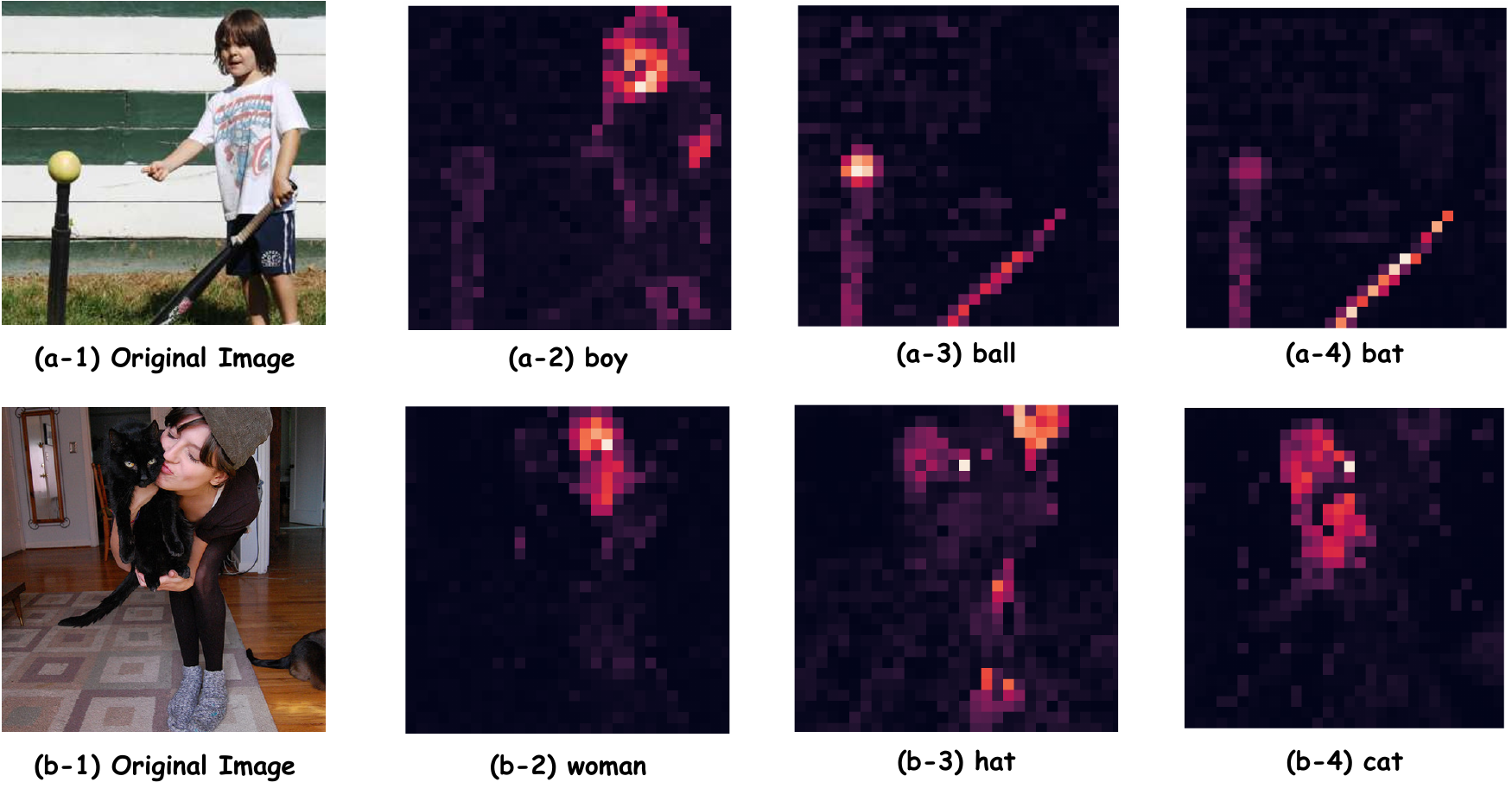}
    \caption{Visualization of the cross-attention maps in the final fusion layer of text tokens. If the color of a attention block is lighter, the cross-attention score is higher.}
    \label{fig:visualization}
\end{figure*}

\subsection{Comparing with End-to-End Baseline Systems}\label{sec:e2e}

ViLT-CAP~\cite{fang2021injecting}, ViTCAP~\cite{fang2021injecting}, E2E-VLP~\cite{xu-etal-2021-e2e} and I-Tuning~\cite{luo2022ituning} share the similar idea as ours to train the image captioning model in an end-to-end manner. Most of them choose to use the pre-trained visual transformer or cross-modal transformer to encode images, but train their language decoders from scratch. Table~\ref{tab:E2E} compares these models with our \textbf{VC-GPT} on MSCOCO test set. One can notice that our framework outperforms all of them across all evaluation metrics. Especially, the ViTCAP model consumes around 4M distinct images during captioning model pre-training, the CIDEr score is still 2.5 lower than ours. These results verify that our \textbf{VC-GPT} is an effective framework for end-to-end image captioning.

\subsection{Comparing with Two-Stage Training Baseline Systems}\label{sec:2stage}

\noindent\textbf{Evaluating on MSCOCO.}\indent Table~\ref{tab:previous_work_comparsion} compares the performance of our \textbf{VC-GPT} and the baseline systems on the MSCOCO test set with different optimization strategies. Notably, our \textbf{VC-GPT} does not require the help of object detectors, which is different from all baseline models.

Comparing our framework and the baseline models without cross-modal pre-training, the overall performance of \textbf{VC-GPT} exceeds all of them by a large margin over different optimization strategies and most evaluation metrics. Considering the large-scale cross-modal pre-trained systems, \textbf{VC-GPT} still can achieve either the best or the second-best performance across all evaluation metrics. Notably, our \textbf{VC-GPT} can achieve a CIDEr score of 127.7 (cross-entropy optimization) on the MSCOCO test set using only 110k images for image captioning pre-training, surpassing the OSCAR model by 4.0 points, which requires more image data.

Beyond this, we consider the systems which are cross-modal pre-trained with small-scale data. Our \textbf{VC-GPT} still outperforms all of them by a large margin (+8.6 for CIDEr, +1.5 for BLEU@4, +1.7 for METEOR and +2.3 for SPICE, of the best model, UNICORN). These results corroborate that our \textbf{VC-GPT} is effective in the large-scale downstream task.\\

\noindent\textbf{Evaluating on Flickr30k.}\indent Flickr30k is a much smaller image captioning benchmark than MSCOCO, with only 29.8k images in the training set, so it is more challenging for models. Table~\ref{tab:flickr} compares the performance of our \textbf{VC-GPT} and the baseline systems. One can notice that the performance gap between our framework and the baseline models becomes even larger. \textbf{VC-GPT} achieves much better overall performance. Especially, our model can achieve a CIDEr score of 79.6, surpassing the XGPT model by more than 8.0 points, while \textbf{VC-GPT} consumes much fewer data during image captioning pre-training. These results reveal that our framework is also effective in the small-scale downstream task.\\

\noindent\textbf{Evaluating on NoCaps.}\indent NoCaps is quite different from MSCOCO and Flickr30k, which majorly focuses on evaluating the generalization ability of image captioning systems. All models are only trained with the training set of MSCOCO and evaluated on three different categories of validation sets. For the in-domain set, the images only contain the same object classes as MSCOCO. For the near-domain set, it contains part-of novel object classes. For out-of-domain set, it only contains novel object classes. Table~\ref{tab:nocaps} compares our \textbf{VC-GPT} and the baseline systems without captioning model pre-training. One can find that \textbf{VC-GPT} achieves the best overall performance. Especially, the overall CIDEr score of OSCAR is even 5 points lower than ours. These results indicate that our end-to-end framework has great generalization ability.

\subsection{Qualitative Evaluation}\label{sec:qualitative}

Apart from the quantitative evaluation in the previous sections, we also evaluate our \textbf{VC-GPT} qualitatively. Figure~\ref{fig:generated_example} presents the image captioning examples of our \textbf{VC-GPT} and OSCAR for the first 4 images in the MSCOCO test set. As can be seen, the generated captions of our \textbf{VC-GPT} depict all images successfully. Moreover, our model successfully identifies the movement of the human in the image. For example, our method can recognize that the girl is blowing the candles on a cake, while OSCAR cannot in the second image. However, our model still fails in some cases, such as recognizing that there are two windows and two sinks in the fourth image. One can notice that such error also exists in OSCAR. The possible reason is that the objects in the image are ambiguous, which is hard for models to recognize.

\subsection{Cross-Attention Map Visualization}\label{sec:attention}

In this section, we visualize the cross-attention maps in our \textbf{VC-GPT} to examine whether it learns the cross-modal alignment implicitly. We randomly choose two images in the MSCOCO dataset and present the cross-attention heatmaps in the final fusion layer of text tokens. Since our model is trained in an auto-regressive manner, the prediction of a token $t_i$ in a sentence is determined by the output hidden states of $t_{i-1}$. Thus, the cross-attention maps of $t_i$ is based on the input token $t_{i-1}$. In addition, there are 12 different attention heads in each fusion layer. We simply average all the cross-attention matrices in the same layer.

As shown in Figure~\ref{fig:visualization}, \textbf{VC-GPT} can correctly attend to the corresponding regions given different tokens. For example, the cross-attention map (see Figure~\ref{fig:visualization}~(a-2) and (b-4)) depicts the contour of the boy and the cat in the images. In addition, the map can also reveal the close relationship between two objects in the image. For example, the attention scores of ``ball'' and ``bat'' are high at the same time in Figure~\ref{fig:visualization} (a-3) and (a-4). These examples reveal that our framework can learn visual grounding implicitly.

\section{Extra Empirical Exploration}

In this section, we empirically explore how different training settings affect our \textbf{VC-GPT}  performance. Specifically, (1) we study whether we can freeze the original parameters of the visual encoder (CLIP) and/or language decoder (GPT2) during pre-training in section~\ref{sec:freeze_or_not}; (2) we explore the influence of different pre-training learning rates and fine-tuning image resolutions in section~\ref{sec:tricks}.

\begin{table}
    \centering
    \caption{Freezing the vision encoder (CLIP-ViT) during pre-training improves the model overall performance on the MSCOCO captioning downstream task. (PT = pre-training)}
    \begin{tabular}{cc|cccc}
        \hline
        \multirow{2}{*}{\textbf{CLIP}} & \multirow{2}{*}{\textbf{GPT2}} & \multicolumn{4}{c}{\textbf{MSCOCO (val)}}\\
        & & \textbf{C} & \textbf{B@4} & \textbf{M} & \textbf{S}\\
        \hline
        \hline
        Freeze & Freeze & 90.8 & 24.3 & 28.7 & \textbf{23.6}\\
        Freeze & PT & \textbf{126.7} & \textbf{37.6} & \textbf{30.0} & \textbf{23.6}\\
        PT & Freeze & 93.7 & 25.4 & 28.4 & 23.2\\
        PT & PT & 125.8 & 37.3 & 29.7 & 22.7\\
        \hline
    \end{tabular}
    \label{tab:freeze}
\end{table}

\subsection{Freezing CLIP-ViT and/or GPT2 during Cross-modal Pre-training?}\label{sec:freeze_or_not}

During the cross-modal pre-training, the model learns how to align the vision and language information in the cross-modal fusion module. For our \textbf{VC-GPT} framework, we choose to freeze the visual encoder during pre-training. In this part, we explore how different freezing strategies during pre-training affect models' performance.

In Table \ref{tab:freeze}, the experimental results reveal that freezing all single-modal pre-trained models or only freezing GPT2 degenerates the overall downstream task performance greatly. Furthermore, only freezing the vision encoder (CLIP-ViT) during pre-training improves overall performance on the downstream task. These results reveal that the freezing strategy of our framework is the best.

\subsection{Image Resolution and Learning Rate}\label{sec:tricks}

\begin{table}
    \centering
    \caption{Comparing different pre-training learning rates for the parameters of GPT2 (G-LR) and the cross-modal fusion module (C-LR). Results are on the MSCOCO captioning downstream task.}
    \begin{tabular}{cc|cccc}
        \hline
        \multirow{2}{*}{\textbf{G-LR}} & \multirow{2}{*}{\textbf{C-LR}} & \multicolumn{4}{c}{\textbf{MSCOCO (val)}}\\
        & & \textbf{C} & \textbf{B@4} & \textbf{M} & \textbf{S}\\
        \hline
        \hline
        1e-5 & 1e-5 & 125.3 & 37.2 & \textbf{30.0} & \textbf{23.6}\\
        3e-5 & 3e-5 & \textbf{126.7} & \textbf{37.6} & \textbf{30.0} & \textbf{23.6}\\
        5e-5 & 5e-5 & 125.0 & 37.1 & 29.7 & \textbf{23.6}\\
        1e-5 & 5e-5 & 124.3 & 36.8 & 29.5 & 23.2\\
        \hline
    \end{tabular}
    \label{tab:LR}
\end{table}

In this part, we intend to empirically study how different learning rates and image resolutions affect our \textbf{VC-GPT} performance. First, \citet{dou2021empirical} indicate that using a larger learning rate for the randomly initialized cross-modal fusion module during pre-training can improve model performance on the downstream tasks. From Table~\ref{tab:LR},\footnote{Since the visual encoder, CLIP, is frozen during VLP, it does not have a learning rate.} we can get a different conclusion that setting the same learning rate for all parameters (except the frozen visual encoder, CLIP) achieves better performance.

\begin{figure}
    \centering
    \includegraphics[height=5.2cm]{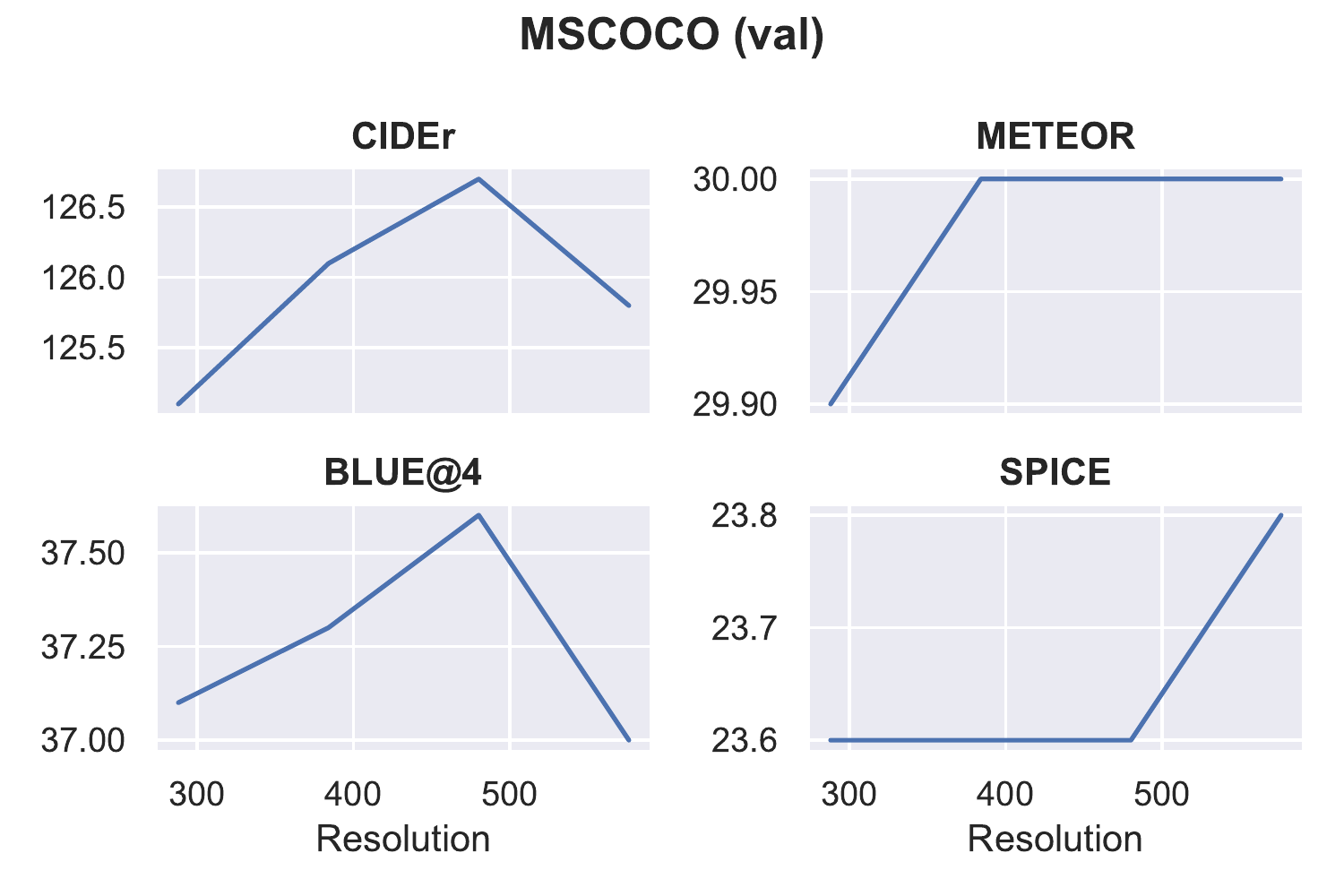}
    \caption{Improving the image resolution during fine-tuning can improve the model performance to a certain extent on the MSCOCO captioning downstream task.}
    \label{fig:resolution}
\end{figure} 

Apart from this, \citet{dou2021empirical} also find that increasing image resolution during fine-tuning can boost performance. A similar finding is also reported in several previous works~\cite{kim2021vilt,yuan2021volo}. We verify such findings in our framework. In Figure \ref{fig:resolution}, we compare different image resolutions during fine-tuning. One can find that increasing the image resolution can improve the model performance to a certain extent, but a larger resolution is not guaranteed to be better. The possible reason is that larger image resolution leads to a larger gap between pre-training and fine-tuning. Note that the resolutions between pre-training and fine-tuning are different, we need to first interpolate the position embeddings of the vision encoder.

\section{Conclusion}

In this paper, we present an end-to-end image captioning framework,  \textbf{Visual Conditioned GPT (VC-GPT)}. Benefiting from our frustratingly simple but highly effective self-ensemble mechanism, our \textbf{VC-GPT} can comprehensively consider both the single- and cross-modal knowledge. Evaluating on three popular image captioning benchmarks, \textbf{VC-GPT} achieves either the best or the second-best performance across all evaluation metrics over extensive baseline systems, while removing the requirement of object detectors.\\


\bibliographystyle{ACM-Reference-Format}
\balance
\bibliography{sample-base}


\end{document}